\documentclass[journal,10pt]{IEEEtran}

\usepackage{amssymb}
\usepackage{amsmath}
\usepackage{cite}
\usepackage{url}
\usepackage{xcolor}
\usepackage{cite,graphicx,amsmath,amssymb}
\usepackage{subfigure}
\usepackage{citesort}
\usepackage{fancyhdr}
\usepackage{mdwmath}
\usepackage{mdwtab}
\usepackage{caption}
\usepackage{amsthm}
\usepackage{algorithm}
\usepackage{algorithmic}
\usepackage{graphicx} 
\usepackage{epstopdf}

\newtheorem{remark}{Remark}
\newtheorem{theorem}{Theorem}

\newtheorem{lemma}{Lemma}

\newtheorem{corollary}{Corollary}

\makeatletter
\def\ScaleIfNeeded{%
\ifdim\Gin@nat@width>\linewidth \linewidth \else \Gin@nat@width
\fi } \makeatother

\begin{document}

\title{Multi-Agent Reinforcement Learning in NOMA-aided UAV Networks for Cellular Offloading}

\author{Ruikang~Zhong,~\IEEEmembership{Student Member,~IEEE,}
Xiao~Liu,~\IEEEmembership{Student Member,~IEEE,}
Yuanwei~Liu,~\IEEEmembership{Senior Member,~IEEE,}
Yue~Chen,~\IEEEmembership{Senior Member,~IEEE,}

\thanks{The authors are with the Queen Mary University of London, London E1
4NS, U.K. (e-mail: r.zhong@qmul.ac.uk; x.liu@qmul.ac.uk; yuanwei.liu@qmul.ac.uk; yue.chen@qmul.ac.uk).}

}

\maketitle
\begin{abstract}
A novel framework is proposed for cellular offloading with the aid of multiple unmanned aerial vehicles (UAVs), while non-orthogonal multiple access (NOMA) technique is employed at each UAV to further improve the spectrum efficiency of the wireless network. The optimization problem of joint three-dimensional (3D) trajectory design and power allocation is formulated for maximizing the throughput. Since ground mobile users are considered as roaming continuously, the UAVs need to be re-deployed timely based on the movement of users. In an effort to solve this pertinent dynamic problem, a K-means based clustering algorithm is first adopted for periodically partitioning users. Afterward, a mutual deep Q-network (MDQN)  algorithm is proposed to jointly determine the optimal 3D trajectory and power allocation of UAVs. In contrast to the conventional DQN algorithm, the MDQN algorithm enables the experience of multi-agent to be input into a shared neural network to shorten the training time with the assistance of state abstraction. Numerical results demonstrate that: 1) the proposed MDQN algorithm is capable of converging under minor constraints and has a faster convergence rate than the conventional DQN algorithm in the multi-agent case; 2) The achievable sum rate of the NOMA enhanced UAV network is $\mathbf{23\%}$ superior to the case of orthogonal multiple access (OMA); 3) By designing the optimal 3D trajectory of UAVs with the aid of the MDON algorithm, the sum rate of the network enjoys $\mathbf{142\%}$ and $\mathbf{56\%}$ gains than that of invoking the circular trajectory and the 2D trajectory, respectively.



\end{abstract}


\section{Introduction}

Owing to the flexible mobility, on-demand deployment, as well as their ability to establish a high probability of line-of-sight (LoS) wireless propagation \cite{UAVov.Zeng}, unmanned aerial vehicles (UAVs) have been invoked as aerial base stations (ABSs) for complementing terrestrial cellular networks in diverse scenarios. On the one hand, UAV-aided wireless networks are practical to be invoked as a backup when the terrestrial cellular networks which rely on ground base stations (GBSs) are paralyzed by natural disasters \cite{Disasters.Zhao}. In these scenarios, UAVs can be employed to displace terrestrial infrastructures for forming temporary communication networks to implement information transfer and disaster relief. On the other hand, UAVs can also be invoked in cellular network offloading scenarios for enhancing connectivity, throughput and coverage of the terrestrial networks \cite{Yuanwei.1}.


Different from the UAV-enabled emergency communication networks in disaster relief scenarios, UAV-aided cellular offloading (UACO) is dedicated to collaborating with the GBS to serve users who cannot be satisfactorily served by the GBS.  It is indisputable that with the continuous development of channel coding, modulation and multiple access technologies since the last century, existing cellular networks have been able to meet the data rate and latency requirements of individual users in the majority of scenarios.  Nevertheless, for some special cases with intensive user density, such as a crowded road or football stadium, the terrestrial infrastructures are not capable of commendable supporting the tele-traffic due to the finite bandwidth and capacity \cite{Crowd}. In this predicament, temporarily deploying a swarm of UAVs and offloading some users from the overloaded GBS to ABSs is regarded as a potential solution to reduce the tele-traffic congestion and improve the quality of service (QoS) \cite{JPA.Liu}. Moreover, the flexibility of UAVs enables the mobile access points to adjust their position to better support various non-ideal user distributions \cite{U.distribution}.

In an effort to tackle the overloading of cellular networks, some techniques which require permanent infrastructures are proposed, such as 5G small hot-spot and micro base stations \cite{SmallBS}. However, compared with the above techniques, UACO is more flexible. UAV fleets are capable of adaptively adjust the fleets size and their positions according to the contemporary user density and distribution.
From an economic perspective, although micro base stations can provide services to crowded venues, as a permanent access point, micro base stations are burdensome to recycle or convert when the service area is desolate. In contrast, UAVs can be recycled and sent to another task after the previous task period, which is more economical and easy to maintain, in line with the appeal for green communications.





\subsection{State-of-the-art}

\subsubsection{UAV-aided Cellular Network Offloading}

Since the deployment of the UAV fleet is acknowledged as a potential scheme, a number of related research contributions were proposed recently. In \cite{UAVCellularMW}, the authors pointed out that the UAVs can be connected with satellites or other kinds of available terrestrial equipment. Thus, UAVs are able to get access to the backhaul network and then provide further connectivity for the congested cellular networks.  The authors of \cite{Edge.Cheng} optimized the trajectory of a single UAV to serve edge users in the adjacent cellular networks. The sum rate of these users was maximized by iteratively optimizing the user scheduling and UAV trajectory. Lyu \textit{et al.} proposed a series of studies on UACO such as \cite{offloading.Lyu} and \cite{Cyclical.Lyu}. The UAV in their model was designed to fly around the GBS following a circular trajectory. The users can alternately get access to the UAV when the UAV is flying over them. The spectrum allocation, user partitioning and UAV trajectory were jointly optimized, but their multiple access scheme do not provide a continuous service to a large number of users and the circular trajectory is not likely to be the optimal solution for the non-ideal user distribution.  Moreover, the authors of \cite{PPP.Turgut} provided a stochastic geometry solution to calculate the signal to interference and noise ratio (SINR) and coverage probability of the UAV-assisted cellular networks. The spectrum trading issue in UACO was studied in \cite{Spectrum.Hu} from a perspective of contract theory.

Some other related contributions on UAV trajectory design and energy efficiency optimization were also impressive though they were not specifically targeted at offloading scenarios, such as \cite{3D.Yaliniz} and \cite{SolarUAV}. The three-dimensional (3-D) placement of a single UAV case was studied in \cite{3D.Yaliniz}.  The authors of \cite{SolarUAV} jointly optimized the 3-D trajectory, power adaptation, and subcarrier allocation for a solar-powered UAV to extend the communication service duration.

\subsubsection{Non-orthogonal Multiple Access (NOMA) in UAV Communications}
Despite its high potential for improving spectral efficiency, research on NOMA enabled UACO is still in an infancy stage. For the single UAV scenario, the author of \cite{JPA.Liu} applied a convex optimization approach to find out the optimum hover position and power allocation for a NOMA enhanced UAV. In \cite{NOMAoff.Kim}, the authors also adopted a circular deployment similar to \cite{Cyclical.Lyu} but employed the NOMA scheme to simultaneously provide service to users who are near and far from the UAV. Both of them figured out that the NOMA technique is able to improve connectivity, reliability and reduce users' outage probability, and the conclusion of \cite{NOMAoff.Kim} also suggested that the optimal trajectory planning and user clustering for the NOMA enhanced UAV worth further explorations.
The authors of \cite{Hover.Song} compared orthogonal and non-orthogonal spectrum utilization in a multiple UAVs engaged case and they maximized the minimum throughput for cellular edge users by jointly optimizing the spectrum allocation, coverage radius, and the number of UAVs.  Furthermore, the studies of the NOMA enhanced UAV up-link \cite{A2GNOMA.Mu} and downlink communications \cite{UAVDLNOMA.Lei} provided evidence that NOMA techniques can improve system reliability and efficiency through productive resource utilization.

\subsubsection{Reinforcement Learning in UAV-aided Wireless Networks}

Reinforcement learning (RL) has demonstrated a vigoroso vitality in optimizing the UAV-aided wireless networks in virtue of its capacity on solving complex, dynamic and non-convex problems \cite{liu2020}. In \cite{TD.Xiao}, an effective Q-learning paradigm was proposed for determining the optimal positions of UAVs to serve ground users. A UAV sensing and relay framework was proposed in \cite{Q.Lingyang}, and the authors employed Q-learning to figure out the optimum sensing positions and trajectories for UAVs. To enlarge the limited state space of the Q-learning model, a combination of Q-learning and neural network (NN), namely deep Q-network (DQN) was proposed \cite{MLOV.MAO}. A DQN based mobility management architecture in UAV-assisted internet of things (IoT) network was presented in \cite{DQN.IOT}.  Recently, the authors of \cite{4ML.UA} introduced the application of various RL algorithms in the UAV relay networks for solving resource management problems, such as the multi-armed bandit learning and actor-critic learning algorithm.


\subsection{Motivations}

Although the aforementioned literature already paved a foundation of solving challenges in the UAV enabled wireless network scenarios and on leveraging NOMA for improving the spectrum-efficiency of networks, the dynamic environment derived from the movement of ground mobile users was ignored in the previous research contributions. Before fully reap the mobility and agility of UAVs, how to design the trajectory of UAVs and resource allocation policy based on the mobility information of users is still challenging in the UACO scenarios.

The main motivation for employing machine learning in UACO is that machine learning constitutes a promising solution for sophisticated problems \cite{Paradigms}.   The two-sides mobility of both UAVs and users make the application of conventional convex optimization more difficult. Additionally, the movement of users and UAVs makes the user association and trajectory design an NP-hard problem. Therefore, facing the aforementioned challenges, current resource management, deployment, mobility management algorithms expose several limitations including but not limited to high complexity, static assumptions, which were pointed out in \cite{MLWN.OV} and \cite{niknam2019federated}.

For such diverse, dynamic and complex problems, machine learning is capable to indicate a near-optimal solution through an experience-based method but not a functional express. The agent accumulates certain experiences through continuous exploration of the current environment and obtains high reward solutions by learning and remembering these fruitful or dreadful experiences \cite{Local.OPT}.



\subsection{Our New Contributions}


In order to remedy these research deficiencies, we put forward the following new contributions:

\begin{itemize}
\item We propose a NOMA-enhanced UACO framework, in which multi-UAV are deployed in 3-D space to complement terrestrial infrastructures. Build on the proposed system model, we formulate the sum rate maximization problem by jointly optimizing the dynamic trajectory of multi-UAV and power allocation policy based on the channel state information of users. Meanwhile, in contrast to the single-cell NOMA system, the dynamic decoding order needs to be determined periodically due to inter-cluster interference. By considering users' mobility, we investigate the sum data rate in the context of the user re-clustering, dynamic decoding order and hierarchical power allocation.

\item We propose a two-step approach to solve the formulated problem. We firstly invoke the upper bounded K-mans algorithm to periodically determine user clusters. Based on the identified user association, a multi-agent MDQN algorithm is proposed to jointly optimize UAVs' 3-D trajectory and power allocation policy to maximize the total throughput. The trajectory derived from the proposed MDQN algorithm not only enables UAVs to establish a desired channel condition with users, but also enables each agent to strive to reduce the interference.

\item Our simulation results demonstrate that: 1. The achievable sum rate of the proposed NOMA framework is superior to the conventional frequency-division multiple access (FDMA) under the same condition. 2. The proposed MDQN algorithm shows a faster convergence rate than the independent DQN scheme and the real-time optimal trajectory design approach outperforms benchmarks, such as the congeneric 2D trajectory and circular deployment. 3. Timely re-clustering is a necessary condition to maintain the optimal sum rate, while invoking dynamic decoding order is proved to have roughly $12\%$ gain in terms of throughput compared to the static decoding order.
\end{itemize}

\subsection{Organization and Notations}

The system model of employing multiple UAVs to offload users for cellular networks is described in Section \uppercase\expandafter{\romannumeral2}, and the problem formulation is illustrated in Section \uppercase\expandafter{\romannumeral3}. Section \uppercase\expandafter{\romannumeral4} details the proposed solution, including the user clustering based on the K-means algorithm and the proposed MDQN algorithm for jointly optimizing the deployment and power allocation. The numerical results are displayed and analyzed in Section \uppercase\expandafter{\romannumeral5}. At last, Section \uppercase\expandafter{\romannumeral6} states our conclusion.

\renewcommand\arraystretch{1.4}
\begin{table}[t!]
 \caption{Notation List}
 \centering
\begin{tabular}{|p{2cm}<{\centering}||p{5cm}<{\centering}|}
 \hline
 Notations & Description \\
 \hline
 $ U $ & Number of UAVs \\
 \hline
 $ K $ & Number of offloaded users \\
 \hline
 $ V_\text{max} $ & User maximum moving speed \\
 \hline
 $ T $ & Offloading service duration \\
 \hline
 $ T_{r} $ & Timing for user re-clustering \\
 \hline
 $ P_\text{LoS}/P_\text{NLoS} $ & Occurrence probability of LoS/NLoS \\
 \hline
 $ L_\text{LoS}/L_\text{NLoS} $ & Pass loss of LoS/NLoS \\
 \hline
 $ l_{k}^u $ & Average pass loss between user $k$ and UAV $u$ \\
 \hline
 $d_{k}^u$ & 3-D distance between user $k$ and UAV $u$ \\
 \hline
 $H_{k}^u$ & Fading coefficient between user $k$ and UAV $u$ \\
 \hline
 $g_{k}^u$ & Channel gain  between user $k$ and UAV $u$ \\
 \hline
 $f_\text{c} $ & Carrier frequency \\
 \hline
 $v_{u,k}$ & Serving indicator \\
 \hline
 $P_{k}^u$ & Allocated power of user $k$ \\
 \hline
 $\sigma$ & AWGN \\
 \hline
 $G_{k}^u$ & Equivalent channel gain \\
 \hline
 $\pi$ & Decoding order \\
 \hline
 $\gamma$ & Signal-to-interference and noise ratio \\
 \hline
 $B$ & Bandwidth \\
 \hline
 $\mathcal{R}$ & Achievable data rate \\
 \hline
 $C_i$ & User cluster $i$ \\
  \hline
 $\mu_i$ & Vector mean of cluster $C_i$ \\
 \hline
 $\eta$ & Maximum UAV load \\
 \hline
 $l_k$ & location of user $k$ \\
 \hline
 $Q()$ & Q value function \\
 \hline
 $S$ & Current state \\
 \hline
 $A$ & Action  \\
 \hline
 $R$ & Reward  \\
 \hline
 $S'$ & Next state  \\
 \hline
 $w$ & Parameters of the neural network\\
 \hline
 $\alpha$ & Learning rate\\
 \hline
 $\beta$ & Discount factor\\
 \hline
 $y$ & Target in MDQN algorithm\\
 \hline
 $J()$ & Loss function \\
 \hline
 $e$ & Size of replay memory \\
 \hline
 $\epsilon$ & Greedy coefficient \\
 \hline
\end{tabular}
\end{table}

\section{System Model}

\subsection{System Description}

\begin{figure*}[t!] 
\centering 
\includegraphics[width=0.7\textwidth]{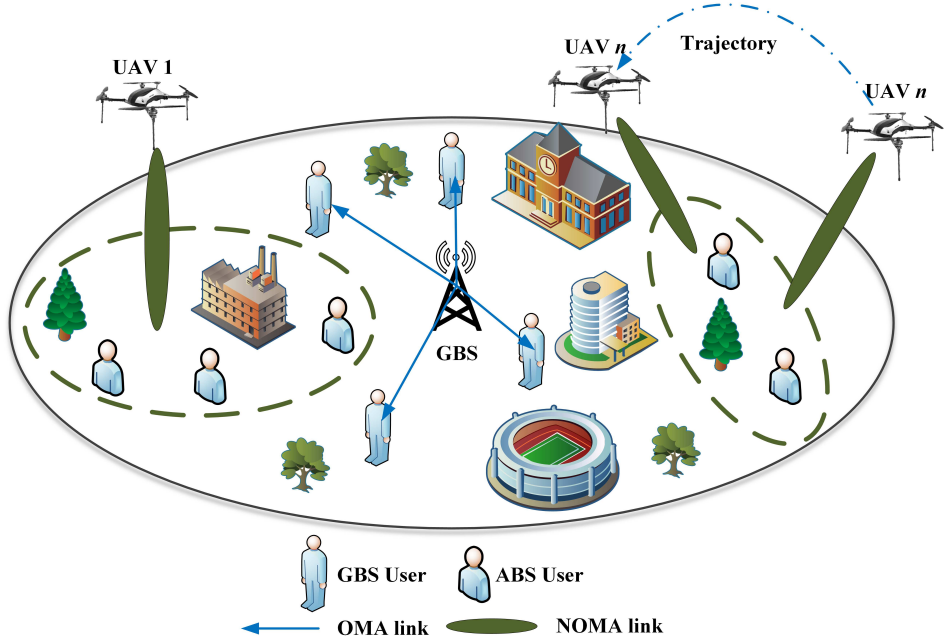} 
\caption{Framework of UAV-aided cellular network with NOMA} 
\label{Fig.main2} 
\end{figure*}

As shown in Fig. \ref{Fig.main2}, we consider an outdoor down-link user-intensive scenario with a central GBS and a number of moving users. Due to the limited user capacity of the GBS, collisions are likely to occur when a number of users request access, which is a nuisance for improving the QoS. In order to provide further connectivity for the overloaded cellular, we propose a multi-UAV-aided cellular offloading framework as a feasible solution. In this scenario, each ABS carried by UAV is equipped with a single antenna and employs the NOMA technique. Different from cellular networks with OMA, NOMA users in the same cell share the same frequency band and suffer from intra-cell interference, which can be subtracted by using successive interference cancellation (SIC) in NOMA networks \cite{NOMA5G.yuanwei}. Moreover, we assume all UAVs are deployed in the cellular utilize the same frequency band, as a result, inter-cell interference has to be considered in this multi-cell network as well. Oppositely, UAVs are assumed to utilize different frequency bands with the GBS to diminish co-channel interference since the GBS has tremendous transmitting power compared to UAVs \cite{GBS.power}. We denote the set of users served by the GBS as $m \in \mathbb{M}=\{1,2,3...M\}$, and the set of users served by the UAVs can be denoted as $k \in \mathbb{K}=\{1,2,3...K\}$, while $\mathbb{M}\bigcap\mathbb{K}=\varnothing$. The users served by UAVs are partitioned into $U$ cells namely user association, where $u \in \mathbb{U}=\{1,2,3...U\}$.  Each user in cell $u$ is only served by UAV $u$, and users in cell $u$ could be clustered into several NOMA clusters. Without loss of generality, in this study, we assume that there is one user cluster associated with each UAV, and in practice, multiple orthogonal resource blocks can be employed by UAVs to serve further user clusters.

\subsection{Mobility Models and User Association}

In this paper, two kinds of user mobility models are invoked, namely random roaming model and directional walking model$\footnote{The proposed solution has extensive applicability to any mobility model, these two typical models are adopted as examples.}$. Users with the random roaming model will move aimlessly, their movement direction and speed are completely random in any discrete time slot $t$. Their moving angle and speed conform to the uniform distribution ${\theta\thicksim U(0,2\pi)}$ and ${V_u \thicksim U(0, V_\text{max})}$. The movement of directional random walking users in each slot $t$ is a vectorial sum of two vectors, a direction vector $\overrightarrow{D_d}$ with fixed direction $\theta=\Theta$, $|\overrightarrow{D_d}|= 4/5 \cdot  V_\text{max}$ and a random vector $\overrightarrow{D_r}$, ${\theta\thicksim U(0,2\pi)}$ and $\thicksim U(0,1/5 \cdot V_\text{max})$.

At the initial time slot of the offloading service, all users offloaded by the GBS are partitioned into several clusters which equal to the number of UAVs according to the users' spatial location. The clustering first ensures that all users will be served and will not be repeatedly served. Secondly, the user clustering which according to the spatial location is helpful to reduce inter-cluster interference since the RL algorithm is likely to drive UAVs to move closer to the served cluster.

\begin{remark}\label{mobility}

Since users are roaming continuously, the initial position of UAVs and user clustering would no longer be optimal at a certain moment, which motivates the re-clustering of users. Re-clustering users in the service area may not necessarily increase the sum data rate but it is a necessary condition for maintaining the optimal data rate.

\end{remark}


As sparked by \textbf{Remark \ref{mobility}}, it is necessary for UAVs to check the user's location and re-cluster the user after a period of time $T_r$. Thus, the total serving period $T$ of UAVs is divided into $T/T_r$ re-clustering time frames.

\subsection{Propagation Model}

The air-to-ground channel model between each UAV and the associated users is provided by the 3GPP specifications Release 15 \cite{3gpp.36.777}. The path loss is depended on line-of-sight (LoS) and non-line-of-sight (NLoS) link states and the design formulas of path loss ${L_{{\text{LoS/NLoS}}}}$ between user $k$ and UAV $u$ can be expressed as \eqref{lk} in the next page, where $h_u(t)$ represents the flight altitude of UAV $u$, $f_c$ represents the carrier frequency, and the 3-D distance from UAV $u$ to user $k$ at time $t$ is denoted as ${d_{{k}}^{u}}(t)$ that

\begin{figure*}
\begin{align}\label{lk}
{{L_{\text{LoS/NLoS}}}(t) = \left\{ {\begin{array}{*{20}{c}}
  {30.9 + \left( {22.25 - 0.5{{\log }_{10}}{h_u}(t)} \right){{\log }_{10}}d_k^u(t) + 20{{\log }_{10}}{f_c},{\kern 1pt} {\kern 1pt} {\kern 1pt} {\kern 1pt} {\kern 1pt} {\text{if}}{\kern 1pt} {\kern 1pt} {\text{LoS}}{\kern 1pt} {\kern 1pt} {\kern 1pt} {\kern 1pt} {\text{link}},} \\
  {\max \left\{ {L_{{\text{LoS}}},32.4 + \left( {43.2 - 7.6{{\log }_{10}}{h_u}(t)} \right){{\log }_{10}}d_k^u(t) + 20{{\log }_{10}}f_c} \right\},{\kern 1pt} {\kern 1pt} {\kern 1pt} {\kern 1pt} {\kern 1pt} {\text{if}}{\kern 1pt} {\kern 1pt} {\text{NLoS}}{\kern 1pt} {\kern 1pt} {\kern 1pt} {\kern 1pt} {\text{link}},}
\end{array}} \right.}
\end{align}
\end{figure*}

\begin{align}\label{dt}
{{d_{{k}}^{u}}(t) = \sqrt {{h_u}^2(t) + {{\left[ {{x_u}(t) - {x_{{k}}^{u}}(t)} \right]}^2} + {{\left[ {{y_u}(t) - {y_{{k}}^{u}}(t)} \right]}^2}} }.
\end{align}

In the propagation model, the probability of LoS is denoted as $P_\text{LoS}$ and described in \eqref{plos} in the next page, where ${d_0} = \max [294.05\cdot{\log _{10}}{h_u}(t) - 432.94, 18]$, while ${p_1} = 233.98\cdot{\log _{10}}{h_u}(t) - 0.95$. Logically, the NLoS probability is ${P_{{\text{NLoS}}}} = 1 - {P_{{\text{LoS}}}}$.  Therefore, the mean path loss between the UAV $u$ and user $k$ can be calculated by \eqref{avgloss}


\begin{figure*}
\begin{align}\label{plos}
{{P_{{\text{LoS}}}(t)} = \left\{ {\begin{array}{*{20}{c}}
  {1,}&{if{\kern 1pt} {\kern 1pt} {\kern 1pt} \sqrt {{{\left( {d_k^u(t)} \right)}^2} - {{\left( {{h_u}(t)} \right)}^2}}  \leqslant {d_0},} \\
  {\frac{{{d_0}}}{{\sqrt {{{\left( {d_k^u(t)} \right)}^2} - {{\left( {{h_u}(t)} \right)}^2}} }} + \exp\left \{\frac{-\sqrt{\left( {d_k^u(t)} \right)^{2}-\left( {{h_u}(t)} \right)^{2}}}{p_1} +\frac{d_0}{p_1} \right \},}&{if{\kern 1pt} {\kern 1pt} {\kern 1pt} \sqrt {{{\left( {d_k^u(t)} \right)}^2} - {{\left( {{h_u}(t)} \right)}^2}}  > {d_0},}
\end{array}} \right.}
\end{align}
\end{figure*}

\begin{align}\label{avgloss}
L_k^u(t)=P_{\text{LoS}} \cdot L_{\text{LoS}} + P_{\text{NLoS}} \cdot L_{\text{NLoS}}.
\end{align}

With the considering of small scale fading, the channel gain from the UAV $u$ to the user $k$ at time $t$ can be calculated as
\begin{align}\label{gt}
{g_k^u(t) = {H_k^u}(t) \cdot {10^{{{ - {L_k}(t)} \mathord{\left/
 {\vphantom {{ - {L_k^u}(t)} {10}}} \right.
 \kern-\nulldelimiterspace} {10}}}}},
\end{align}
where ${H_k^u}(t)$ represents the fading coefficient\cite{offloading.Lyu} between UAV $u$ and user $k$ .


\subsection{Signal Model}

Denote ${v_{u,k}}$ as the serving indicator. ${v_{u,k}}=1$ represents the UAV $u$ is serving the user $k$, ${v_{u,k}}=0$ if otherwise. Thus, the superposition  transmitting signal $x^u(t)$ of the UAV $u$ can be calculated as~\cite{cui.signal}

\begin{align}\label{xnu}
{x^u(t) = \sum\limits_{k = 1}^K {{v_{u,k}}(t)\sqrt {P_{k}^u(t)} } x_{k}^u(t) },
\end{align}
where $x_{k}^u(t)$ is the transmitting signal from UAV $u$ to user $k$, $P_{k}^u(t)$ denotes the allocated power of user $k$. As a consequence of Equation \eqref{gt} and \eqref{xnu}, the received signals at user $k$ is

\begin{align}\label{ynuk}
{y_{k}^u(t) = g_{k}^u(t)x^u(t) + {I_\text{inter}}_{k}^u(t) + {I_\text{intra}}_{k}^u(t) + \sigma _{k}^u(t) },
\end{align}
where $\sigma _{k}^u(t)$ represents the additive white Gaussian noise (AWGN). ${I_\text{inter}}_{k}^u(t)$ is the accumulative inter-cluster interference to user $k$ from other UAVs except UAV $u$ and ${I_\text{intra}}_{k}^u(t)$ represents intra-cluster interference.

The composition of ${I_\text{inter}}_{k}^u(t)$ can be expressed as


\begin{align}\label{inuk}
{{I_\text{inter}}_{k}^u(t) = \sum\limits_{s = 1,s \ne u}^U {g_{k}^s(t)\sqrt {P^s(t)} x^s(t)} },
\end{align}
where $g_{k}^s(t)$ denotes channel gain between UAV $s\neq u$ and user $k$, $P^s(t)$ represents the total power consumption of UAV $s\neq u$, whcih is given by

\begin{align}\label{pns}
{P^s(t) = \sum\limits_{k = 1}^K {{v_{u,k}}(t)P_{k}^s(t)} }.
\end{align}

The precondition of determining ${I_\text{intra}}_{k}^u(t)$ is to find out the optimal decoding order to guarantee the successful SIC, and then SIC is capable to remove some of the intra-cluster interference at the receiver side \cite{SIC}. In this case, a dynamic decoding order has to be considered owing to the fact that the channel gain and inter-cluster interference of each user is always changing by the movement. The auxiliary term $G_{k}^u(t)$ shown in \eqref{gnku} is interjected as a criterion for determining the decoding order, and $G_{k}^u(t)$ can be regarded as the equivalent channel gain.

\begin{align}\label{gnku}
{G_{k}^u(t) = \frac{{{v_{u,k}}(t)g_{k}^u(t)}}{{\sum\nolimits_{s = 1,s \ne u}^T {g_{k}^s(t)P^s(t) + {{\sigma _{k}^u(t)} ^2}} }}}.
\end{align}


Since the position of both UAVs and users is time-varying, the dynamic decoding order has to be determined at each time slot to guarantee the successful SIC. Consider a NOMA cluster with user $j$ and user $k$ associated with UAV $u$, and their equivalent channel gains can be noted as $G_{k}^u(t), G_{j}^u(t)$, respectively. Then the condition of user $k$ to remove the signal of user $j$ by SIC is that
\begin{align}\label{EG}
G_{k}^u(t) \ge G_{j}^u(t),
\end{align}
which can be derived from \cite{cui.signal}.  Inequation \eqref{EG} suggests that SIC is supposed to be implement at the receiver with stronger equivalent channel gains.

Extending the above principle to a NOMA cluster $u$ with $K^u$ users, we can figure out the decoding order according to the equivalent channel gain, noted as $G_{\pi (1)}^u(t) \le G_{\pi (2)}^u(t) \le  \cdots  \le G_{\pi (K^u)}^u(t)$, where $\pi (k)$ denotes the decoding order of user $k$. According to the SIC principle, the user $\pi(k)$ decodes and successively subtracts the signals for all the $\pi (k-1)$ users, and then decode the desired signal. With this principle, since the signals for $(k-1)$ users are removed, the intra-cluster interference ${I_\text{intra}}_{\pi(k)}$ and desired signal for user $\pi(k)$ be calculated as
\begin{align}\label{Iintra}
{{I_\text{intra}}_{\pi(k)} = \sum\limits_{i=k+1}^{K^u}{v_{u,\pi (i)}}(t)g_{\pi (i)}^u(t)P_{\pi (i)}^u(t)x_{\pi (i)}^u(t)}.
\end{align}

\begin{align}\label{Desired}
{S_{\pi(k)} = {v_{u,\pi (k)}}(t)g_{\pi (k)}^u(t)P_{\pi (k)}^u(t)x_{\pi (k)}^u(t)}.
\end{align}

Build on Equation \eqref{inuk} \eqref{Iintra} and \eqref{Desired}, the SINR for the $k$-th decoded user is given by \eqref{rnpiku}.

\begin{figure*}\label{rnpiku}
\begin{align}\label{rnpiku}
{\gamma _{\pi (k)}^u(t) = \frac{{{v_{u,\pi (i)}}(t)g_{\pi (k)}^u(t)P_{\pi (k)}^u(t)}}{{\sum\nolimits_{i = k + 1}^{ K^u } {{v_{u,\pi (i)}}(t)g_{\pi (i)}^u(t)P_{\pi (i)}^u(t) + \sum\nolimits_{s = 1,s \ne u}^U {g_{k}^s(t)P^s(t) + {{\sigma _{k}^u(t)} ^2}} } }}}.
\end{align}
\end{figure*}

Then the data rate of user $k$ connected with UAV $u$ can be calculated as

\begin{align}
\mathcal{R}_{\pi (k)}^u(t) = {B}\log 2\left( {1 + \gamma _{\pi (k)}^u(t)} \right),
\end{align}
where $B$ represents bandwidth of UAV $u$. Hence, the overall data rate of UAV service at time $t$ can be expressed as
\begin{align}\label{Rnpiku}
{{\mathcal{R}(t)} = \sum\limits_{u = 1}^U \sum\limits_{k = 1}^K{{\mathcal{R}_{\pi (k)}^u(t)} } }.
\end{align}

Therefore, the throughput during the serving period is

\begin{align}\label{Ravg}
\mathcal{R} = \sum_{t=0}^{T}\mathcal{R}(t).
\end{align}

%
%
%

\section{Problem Formulation}

Intending to maximize the total throughput, we optimize the trajectory and power allocation policy of UAVs, subject to the maximum power constraint, spacial constraints, and the QoS constraint. The problem is formulated in \eqref{OPP}.  $\mathbf{H} = \{h_u(t),0\leq u\leq U, 0\leq t\leq T\}$ represents the positions of UAVs during service time $0\leq t\leq T$ , and the velocity of UAVs is assumed as fixed. The transmitting power of each UAV is denoted as $P_u$, and the power allocation policy is denoted as $\mathbf{P} = \{p_k(t), 0\leq t\leq T, k \in {\mathbb K}\}$. Finally, user-UAV indicator $\mathbf{V} = \{v_{u,k}(t), t = T_r, u\in {\mathbb U}, k\in {\mathbb K}\}$ is used to represent the user association. Hence, the optimization problem can be formulated as



\begin{subequations}
\begin{align}\label{OPP}
\max_{\mathbf{H,V,P}} \quad &\mathcal{R} = \sum_{t=0}^{T}\mathcal{R}(t), \\
\textrm{s.t.}
&{h_{\min }} \le {h_u(t)} \le {h_{\max }},\forall u,\forall t,\notag \\
&{x_{\min }} \le {x_u(t)} \le {x_{\max }},\forall u,\forall t,\notag\\
&{y_{\min }} \le {y_u(t)} \le {y_{\max }},\forall u,\forall t \label{OPPB},\\
&{\sum_{u=1}^{N} v_{u,k}=1 }, \label{OPPD}\\
&\sum\limits_{k \in {\mathbb K}} { {{v_{u,k}}(t)P_k^u \le {P_u}} }, \forall t,\forall u,\forall k,\label{OPPE}\\
&G_{\pi (k)}^u \ge G_{\pi (j)}^u,k > j,\forall (k,j),\forall t,\forall u,\label{OPPF}\\
&R_k(t) \geq R_{\text{QoS}}, \forall k, \forall t, \label{OPPG}
\end{align}
\end{subequations}
where \eqref{OPPB} indicates the constraints for UAVs' 3-D position, which has to be in the airspace above the service area within achievable height range to avoid the collision between UAVs or interfere other communication equipment outside the offloading cellular.  Constraint \eqref{OPPD} ensures each user $u\in \mathbb U$ only be served by one UAV.  Constraint \eqref{OPPE} denotes the transmitting power constraint to guarantee the power consumption of each UAV never beyond the upper transmitting power bound.  Constraint \eqref{OPPF} represents the decoding order for successful SIC.  Constraint \eqref{OPPG} formulates the rate constraint in terms of fairness of users. Since the problem category of \eqref{OPP} was proved to be NP-hard in \cite{NP.HARD}, and the formulated problem is with highly dynamic due to the movement of UAVs and users, it is challenging for the conventional convex-optimization algorithms to solve the formulated problem. Thus, the RL-based algorithm, which can interact with the environment and learn from the interactive experience, is invoked in this paper.

\section{Proposed Solutions}

This section introduces the proposed solution which contains two parts, user clustering as well as the optimization for trajectory and power allocation. The first subsection introduces the spatial user association by adopting K-means clustering with an upper bound limitation of the cluster members. The second subsection presents a multi-agent MDQN design for the UACO scenario. At the end of this section, complexity analyses for the proposed algorithm are provided.

\subsection{K-means Based User Clustering}

K-Means algorithm is a heuristic algorithm, which has been proved to achieve favorable performance for user clustering in wireless communication \cite{cui.clustering}. The K-means algorithm is designated since the low calculation complexity of the K-means clustering can timely find out clustering results and prevent service interruption when re-clustering is carried out. The secondary reason is that the K-means algorithm does not need any prior knowledge for training.


In this application, the position $\mathbb{L}=\{ l_1, l_2\ldots l_K\}$ of users $k \in \mathbb{K}$ is input as the observation set. The user set has to be partitioned into $U$ clusters according to the spatial distance between the user samples to make the users within the clusters connected as closely as possible. To achieve that, an conventional K-means algorithm first applied to find the cluster partition $(C_1, C_2 \ldots C_U)$ with the minimum SSE in \eqref{SSE}. The algorithm initially chooses $K$ random users as the initial centroid of each cluster, then assigns each user to the nearest cluster and recalculates the centroid of each cluster. The algorithm iterates this process until the centroids of all clusters no longer change.

\begin{align}\label{SSE}
SSE = \sum_{k=1}^{K}\sum_{l\in C_i} \| l - \mu_i\|^2,
\end{align}
where $\mu_i$ is the vector mean of cluster $C_i$, also known as centroid as shown as \eqref{centroid}

\begin{align}\label{centroid}
\mu_i = \frac{1}{|C_i|} \sum_{l\in C_i} l.
\end{align}

It is worth noting that the user capacity of a single UAV is finite, which is not considered in the conventional K-means clustering. Hence, a necessary supplement is imposed when any cluster has the number of users out of the UAV's load ability. Supposing there are excess elements in a user cluster, the farthest user will be removed and assigned to another closest cluster. This operation will be repeated until all clusters with a legal number of users. The flowchart and the formula of the entire algorithm are clearly given in  \textbf{Algorithm \ref{K-means}}. 

\begin{algorithm}
\caption{K-means based user clustering algorithm}
\label{K-means}
\begin{algorithmic}[1]

        \STATE Initialize desired cluster number $(C_1\ldots C_U)$, the maximum number of iterations $N$, maximum UAV load $\eta$
        \STATE Input users location $\mathbb{L}=\{ l_1, l_2\ldots l_K\}$ as observation set
        \STATE Randomly select $U$ samples in $\mathbb{L}$ as initial centroid $(\mu_1\ldots \mu_U)$
        \FOR{$n = 1,2 \ldots N$}
        \FOR{$l \in \mathbb{L}$}
		\STATE Calculate $d_{ku} = \| l_k - \mu_u\|$
        \STATE Allocate $l_k$ to $C_u$ with minimum $d_{ku}$
        \STATE Update $\mu_u$ according to \eqref{centroid}
        \ENDFOR
        \IF {$\mu_u(n) = \mu_u(n-1)$}
        \STATE End loop
        \ENDIF
        \ENDFOR
        \WHILE{$|C_u| > \eta$}
        \STATE Remove $l_k$ with $max (d_{ku})$ from $C_u$
        \STATE Add $l_k$ to $C_i \neq C_u$ with minimum $d_{ki}$
        \ENDWHILE
        \STATE output $(C_1\ldots C_U)$

\end{algorithmic}
\end{algorithm}

\subsection{MDQN Algorithm for Deployments and Power Allocation}

\begin{figure*}[t!] 
\centering 
\includegraphics[width=0.9\textwidth]{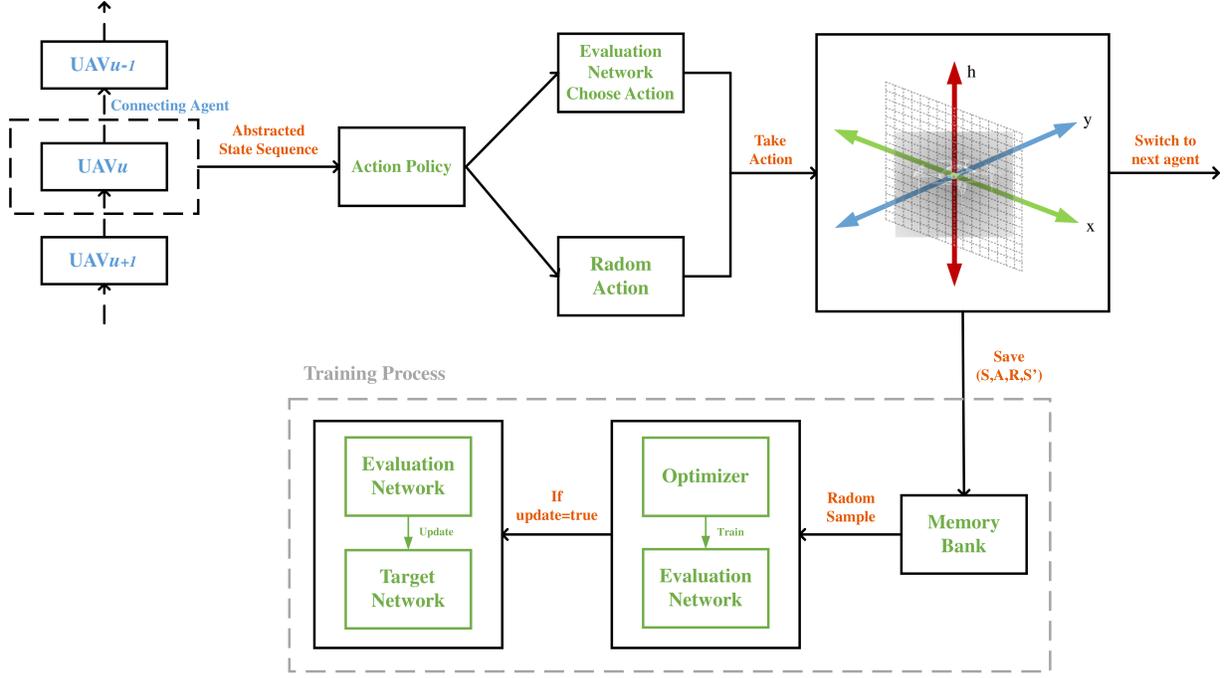} 
\caption{MDQN based trajectory design and power allocation flow diagram} 
\label{Fig.DQN} 
\end{figure*}

Since a plurality of UAVs is deployed in the cellular offloading scenarios, a multi-agent MDQN algorithm is designed to jointly optimize UAV trajectories and power allocation. These UAVs are considered as independent agents to choose their actions, but multiple UAVs are permitted to connect with the same NN during the training process with the assistance of state abstraction. In this paradigm, although the experience of each agent is different, it can be reorganized into a standard form and then these experiences can be used to train a mutual NN. It can also be considered that the standardized experience of each agent can also be indirectly obtained by other agents via the shared NN. Thus, the training time is compressed and the lengthiness training problem of the conventional DQN paradigm is alleviated. It is worth to note that the MDQN paradigm only requires data exchange during the training process. When UAV provides services, the parameters of NN will be copied to each agent. The agent can choose to update its own NN locally or only use NN for prediction without updating parameters. Hence, the MDQN paradigm does not require additional communication between UAVs during service, and the computational complexity of the state abstraction is at a negligible level, which is analyzed in the next subsection. Moreover, the proposed algorithm can be considered as an online solution since the MDQN algorithm does not give a predetermined trajectory or power allocation policy.

\subsubsection{State, Action Value and Reward}

DQN algorithm is a value-based reinforcement learning method, which chooses actions in discrete action spaces when confronting a situation, namely, state ($S$).  The quality of the decision, which called the reward ($R$), is evaluated by Q value. The agent chooses actions according to the Q value which is estimated by the NN to maximum long term reward.

DQN algorithm learns experience by repeating training scenes, and each repetition is called an episode. In each episode, the artificial intelligence entity, namely agent, will recognize the current state $S$, and then select and implement the action ($A$) based on action policy. The implementation of action leads to a change in the environment, turning state $S$ to the next state $S'$ following the Markov process \cite{liu2019MDP}, and the agent will save $S, A, R$ and $S'$ as an experience to train the NN.  In order to maximize the long term reward, the DQN algorithm always chooses the action with maximal Q value. The Q value can be calculated according to the action value function, also known as the Bellman equation, expressed in \eqref{Bellman}, where $\alpha$ represent learning rate ( $ 0 < \alpha < 1$), and $\beta$ is discount factor ( $ 0 < \beta < 1$ ). In order to reduce the calculation complexity of the Q value, in the DQN algorithm, the action value function is approximated by the NN \cite{chen2020bellmen}.

\begin{align}\label{Bellman}
Q(S,A) \leftarrow Q(S,A) + \alpha[R + \beta \max Q(S',A') - Q(S,A)].
\end{align}
A number of prior studies revealed that NN can be used to fit complex functional relationships and the order of the relationship function will not lead to sharp increases in the complexity of the neural network structure. Nevertheless, as an online algorithm, the training process of the NN is still a heavy burden for hardware. In order to further reduce the training complexity, we proposed the reformative MDQN paradigm based on the conventional DQN algorithm.

\subsubsection{Schema Design of the MDQN Algorithm}

The schematic diagram of the MDQN engaged UAV networks is depicted in Fig.~\ref{Fig.DQN}. In the MDQN model, agents need to connect to the NN accordingly. The connecting UAVs firstly input the abstracted state information into the evaluation network to determine the optimal action. Afterward, rewards are calculated and actions will be executed in the environment. After all UAVs finish carrying out actions, the data rate in this time slot will be calculated. The detailed algorithm flow is listed in \textbf{Algorithm \ref{DQN TDPA}}.

This MDQN algorithm establishes a target network with identical structures of evaluation network for training since the single network design will make NN become destabilized by falling into feedback loops between the target and estimated Q values \cite{mnih2013playing}. Therefore, a delayed target network is employed to avoid the estimated values spiral out of control. The parameters of the evaluation network are updated with each training step and the evaluation network is used to estimate the Q value of actions. The parameters of the target network are relatively stable. After a number of steps, the target network will update the parameters to the same as the evaluation network.

\begin{figure*} \centering
\subfigure[MDQN neural network connection] {
 \label{fig:NNa}
\includegraphics[width=0.6\textwidth]{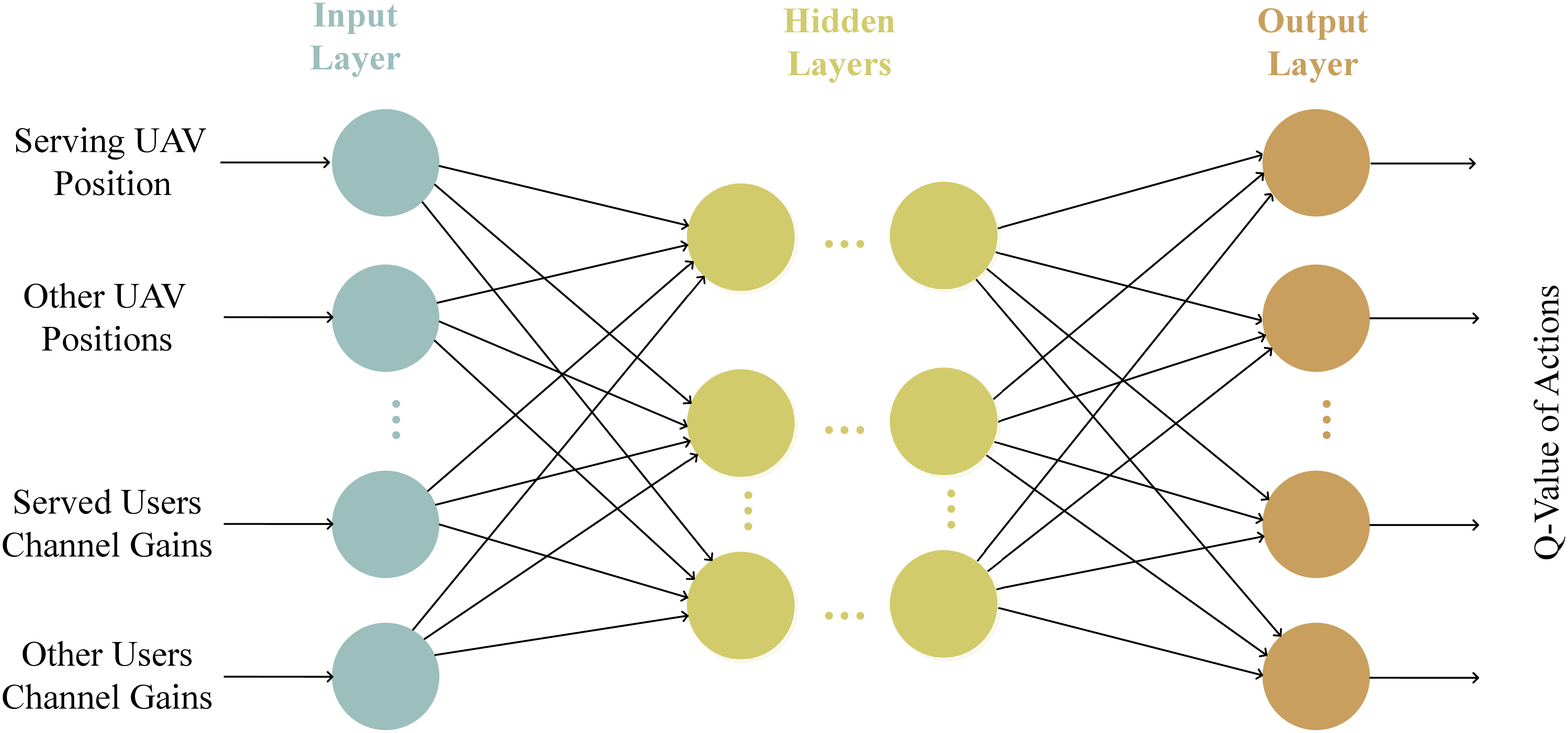}
}
\subfigure[Conventional DQN neural network connection] {
\label{fig:NNb}
\includegraphics[width=0.6\textwidth]{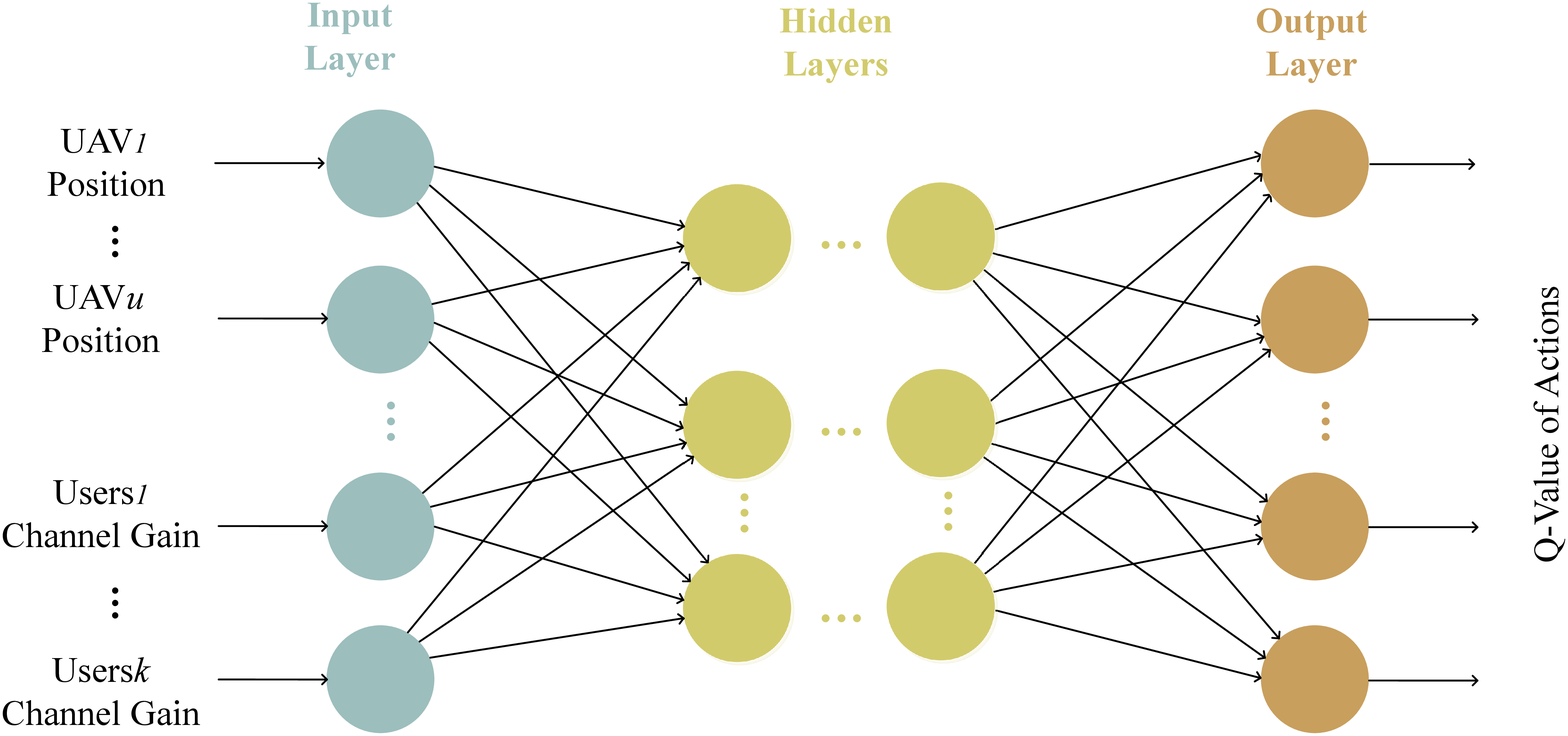}
}

\caption{ Neural network input connection comparison between MDQN and DQN algorithm}
\label{Fig.NN}
\end{figure*}

\begin{algorithm}
\caption{MDQN algorithm for deployments and power allocation}
\label{DQN TDPA}
\begin{algorithmic}[1]

		\FOR{each episode}
        \STATE Initialize initial positions of UAVs and users
        \STATE Initialize the evaluation network $w_e$ and the target network with random parameter $w_t$
        \STATE Update $\epsilon$ in action policy
        \FOR{each step $ t_0 \leq t \leq t_0 + T_r$}
		\FOR{each UAV}
        \STATE Calculate $G_k^u, k\in \mathbb{K}$
        \STATE Generate state abstraction array $S$
        \STATE Choose $A$ according to action policy and $Q(S,A,w_e)$
        \STATE Take action $A$, observe $R$ and $S'$
        \STATE Store $e = (S,A,R,S')$
        \STATE Sample random pair of $e$ from memory

        \STATE Calculate target $y = R + \beta \max Q(S',A',w_t)$
        \STATE Train parameter $w_e$ with a gradient descent step $(y - Q(S,A,w_e))^2$
        \IF{update = true}
        \STATE $w_t \leftarrow w_e$
        \ENDIF

        \STATE $S \leftarrow S'$
        \ENDFOR
        \STATE Users move
		\ENDFOR
        \ENDFOR
\end{algorithmic}
\end{algorithm}

\subsubsection{State Abstraction}

Since the MDQN model needs to calculate the Q value of actions according to the input state information $S$, which is formed by positions of UAVs and user channel gain in this model. The 3-D positions of UAVs are considered to be the current deployment state of UAV, but the position of users $L_{u}$ is difficult to be captured at all times. Thus, the channel gain $g_{k}^u$ which known by UAVs and described in \eqref{gnku} is used to describe the state of the user as the basis for power allocation $\footnote{Please note that users' position can also be used as the input parameter, but generally, the channel gain is more easily estimated by the UAV.}$.

In order for multiple UAVs to share the NN, the state information from each UAV have to be abstracted and shuffled into a standard array before the state information is entered into NN. The principles of shuffling are illustrated in Fig.~\ref{Fig.NN}, the UAV currently connected to the neural network needs to latch its input neurons. For example, when UAV1 is connecting to the NN, the location information of UAV1 is input into the first neuron, and when UAV2 is activated, the coordinates of UAV2 have to be input to the same neuron node as well. In other words, the connecting UAV and the users served by the connecting UAV must use the specified input neural nodes. By the feat of this design, the neural network approximates the logical relationship between the interferer and the victim and this logic is universal for all UAV with equivalent equipment.

Moreover, since $L_{u}$ and $g_{k}^u$ have different dimensions and excessively divergent magnitude, in order for the MDQN algorithm to efficiently process these mixed data, scalarization and scaling is suggested to be taken. The 3D coordinate $L_{u}$ has to be split into three scalars before inputting into NN. The input state array $S$ can be expressed as

\begin{align}\label{S}
S = \{L_{u}(t),L_{s}(t),g_{k}^u(t),g_{k}^s(t)\}, u,s \in \mathbb{U}, s\neq u, k \in \mathbb{K},
\end{align}
where $L_{u}(t)$ is denote the 3D coordinate of the connecting agent and $L_{s}(t)$ denote coordinates of other agents, which are considered as sources of inter-cluster interferences. Analogously, $g_{k}^u(t)$ and $g_{k}^s(t)$ represent the channel gain of associated users and the channel gain of users associated with other UAVs, respectively.

\begin{remark}\label{Expshare}

State abstraction makes it possible for multiple agents to jointly train an NN. Compared to the approach that multi-agent train NN independently, the proposed approach can significantly increase the convergence rate. The three-phase in state abstraction, shuffling, scalarization and scaling are essential, otherwise, NN may not able to converge.

\end{remark}

\subsubsection{Action Space}

The action space contains two subsets, UAV movement actions and power allocation policies for the next step. Since continuous action space will cause infinite complexity for choosing actions, discretization is considered necessary in the action space design. Therefore, UAV is set to perform some representative flight maneuvers, and the transmission power for each user is also preset to several fixed gears for UAV to choose from. Meanwhile, the necessary premise of an experience sharing is that all agents are required to have the same action space so that the shared experience is beneficial and correct. Therefore, specifically, all agents have the same following action space:

\begin{itemize}
\item Movement action space: UAV is authorized to choose an action from seven flight actions, namely, \{horizontal left, horizontal right, horizontal forward, horizontal backward, vertical upward, vertical downward, hover\}. Corresponding to \eqref{OPPB}, when the UAV is flying out of the border, then this action is considered to be invalid and the hover will be performed by default.
\item Power allocation action space: Since the MDQN model outputs discrete actions, the power distribution for each user is preset to multiple gears ${P_1,P_2 \ldots P_p}$. The agent will select and maintain a power gear for each associated user until the next action.
\end{itemize}

\subsubsection{Neural Network Training}
In order for the NN to accurately estimate the Q value, the NN needs to be trained by a number of samples. To reduce the correlation of sampling, we use memory replay technology in the training of MDQN. At the early stage in training, agents take random implementation actions, and store the experiences in a memory bank. The experiences contain the information of $S$, $R$, $A$ and $S'$ and it will be used as training samples. The NN with parameter $w$ can be trained by minimizing the loss function \eqref{Loss}, where $y$ denotes target and the loss function $J()$ is changeable depending on characteristics of optimization problems.

\begin{align}\label{Y}
y = R + \beta \max Q(S',A'),
\end{align}
\begin{align}\label{Loss}
J(w) = J[(y - Q(S,A,w))^2].
\end{align}

After obtaining some knowledge, the agent probabilistically chooses random actions (exploration) or the optimal action (exploitation) according to the action policy, where the agent makes decisions according to action policy and the estimation from NN.

\subsubsection{Action Policy}

During the training process, a $\epsilon-greedy$ action policy with a decreasing $\epsilon$ is adopted to guide the agent to choose actions as shown in Fig.~\ref{Fig.DQN}. This policy makes the agent have the probability of $\epsilon$ to choose the exploration, and logically the agent has a probability of $1-\epsilon$ to choose the exploitation. Mathematically, it can be expressed as

\begin{equation}
A=
\begin{cases}
random\ action, &  \epsilon, \\
{argmax}_A Q(S,A,w_e),&  1-\epsilon.
\end{cases}
\end{equation}

\subsubsection{Reward Function}
As mentioned in equation \eqref{OPP}, the objective function is maximizing the total throughput under the constraint of guaranteeing the user fairness \eqref{OPPG}, so the reward function is designed as

\begin{align}\label{Reward}
R_{u} =\frac{\mathcal{R}(t)}{2^\lambda}, u \in U,
\end{align}
where $R(t)$ is the sum data rate and $\lambda$ is the penalty coefficient. Since multiple UAVs are in collaboration, in order to maximize the throughput of the system, the reward of each agent has to be determined by the sum rate rather than its own data rate. In this setting, each UAV will take the most beneficial action to improve system performance based on the State information it now understands. The penalty coefficient is introduced to enforce the agent to guarantee the QoS of each user as much as possible. The penalty coefficient increases when the agent chooses a route that violates the QoS requirement. $\lambda$ stops raising when increasing it cannot reduce the number of steps that do not meet the QoS requirements. This situation means that in those steps, no action can meet the QoS of a certain user.


\subsection{Computational Complexity Analysis}

This subsection discusses the computational complexity of the proposed algorithm in the case of multiple UAVs.

\begin{itemize}
\item The complexity of K-means based clustering algorithm: The complexity of using K-means algorithm to obtain the clustering for $U$ UAVs and $K$ users is $\mathcal{O}(2 N \cdot U \cdot K)$, where $N$ represents maximum iteration number. The cluster member number checking and correcting complexity is $\mathcal{O}(2\cdot U \cdot K)$, while the association step costs $U^2$. Thus, the approximate total complexity is $\mathcal{O}(N \cdot U \cdot K)$.

\item The complexity of action selection in MDQN algorithm: The complexity for the MDQN model to make a single decision is $\mathcal{O}(|A| \cdot n + |S| \cdot n + |A|)$, where $n$ is the number of nodes in hidden layers and $|S|$ represent the size of $S$. Since Q-value of each action needs to be calculate, the complexity of each step is $\mathcal{O}(|A| \cdot  (|A| \cdot n + |S| \cdot n + |A|)) $.  The complexity of state abstraction and action decoding for one step is $\mathcal{O}(2U \cdot K)$ which is negligible compared with calculating Q-value. Therefore, in a task with a step number $t$, the total complexity is $\mathcal{O}( t\cdot U\cdot |A| \cdot (|A| +|S|)\cdot n) $.

\item The complexity of training step in DQN algorithm: Assume that the number of episodes is $E$ and the batch size is B, the complexity caused by the action selection in DQN algorithm during training is $\mathcal{O}( E \cdot B \cdot t\cdot U \cdot |A| \cdot(|A| +|S|)\cdot n))$. The complexity of training NN is $\mathcal{O}( E \cdot B \cdot t\cdot U\cdot n))$. Therefore the total complicity is still $\mathcal{O}( E \cdot B \cdot t\cdot U \cdot |A| \cdot(|A| +|S|)\cdot n))$ approximately.

\end{itemize}

\begin{remark}
The computational complexity of state abstraction is negligible comparing with action selection. We can obtain the relationship $|S|\sim 3U+K $ according to \eqref{S} and $|A|\sim 7U+|P|K $ as suggested by action space definition, where $|P|$ represents the gears number of power. As a result, it can be easily proved that the complexity of state abstraction is negligible compared with NN calculating Q-value. Therefore, the proposed MDQN algorithm does not cause a significant increase in complexity, compared to conventional DQN algorithm.
\end{remark}

\section{Numerical Results and Analysis}

This section provides numerical results to validate the effectiveness of the proposed approaches and evaluate the gain of each component in the proposed approaches.  In the simulation, users are randomly distributed in the service area and UAVs are deployed near the boundary of the cellular with a height of 100 meters at the initial time. The employed neural network is with 3 layers and a 40-nodes hidden layer. The activation function is rectified linear units and mean squared error is chosen as the loss function. An Adam optimizer is applied for training the NN.  The $\epsilon$ for greedy action policy is set to linear decreasing from 0.9 to 0. The rest of the default simulation parameters are listed in Table \ref{SP}. In the case of no special explanation, the simulation is with default parameters.

\begin{table}[t!]
 \caption{Simulation Parameters}\label{SP}
 \centering
 \begin{tabular}{|p{1.7cm}<{\centering}||p{3.5cm}<{\centering}||p{1.7cm}<{\centering}|}
  \hline
  Parameter & Description & Value \\
  \hline
  $f_\text{c}$ & Carrier frequency & 2GHz\\
  \hline
        $U$ & Number of UAVs & 3  \\
        \hline
        $B$ & bandwidth for each RB & 15 kHz \\
        \hline
        $K$ & Number of offloaded users & 6 \\
        \hline
        $V_\text{max}$ & User maximum moving speed & 0.5 m/s \\
        \hline
        $V$ & UAV speed & 5m/s \\
        \hline
        $P_\text{max}$ & \multicolumn{1}{p{3.5cm}||}{Maximum total transmitting power of UAVs} & 29 dBm \\
        \hline
        $T_{r}$ & Timing for user re-clustering & 60 s \\
        \hline
        $h_\text{max}$ & Maximum UAV altitude & 150 m \\
        \hline
        $h_\text{min}$ & Minimum UAV altitude & 20 m \\
        \hline
        $x_\text{min},y_\text{min}$ & Service area boundary & 0 m \\
        \hline
        $x_\text{max},y_\text{max}$ & Service area boundary & 500 m \\
        \hline
        $R_\text{QoS}$& QoS require & 0.15kb/s \\
        \hline
        $\sigma$& AWGN power & -100 dBm/Hz \\
        \hline

        $\alpha$ & Learning rate & 0.001\\
        \hline
        $\beta$ & Discount factor & 1\\
        \hline
        $e$ & Size of replay memory & 10000 samples\\
        \hline
        $\upsilon$ & Update frequency & 600-2000 \\
        \hline
        $\omega$ & Batch size & 128 samples \\
        \hline
        $\epsilon$ & Greedy coefficient& 0 - 0.9\\
        \hline

 \end{tabular}
\end{table}

\begin{figure}[t!] 
\centering 
\includegraphics[width=0.5\textwidth]{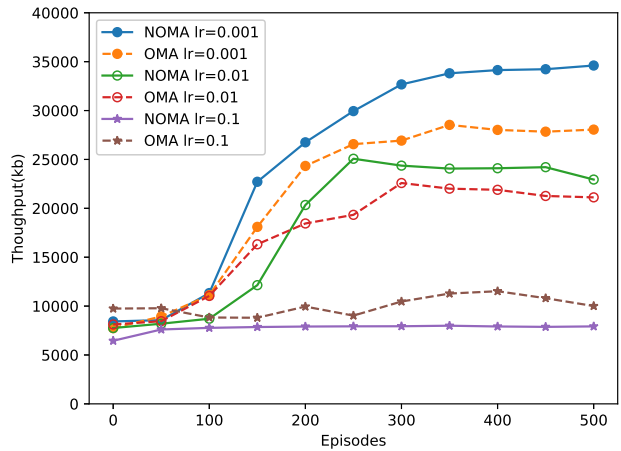} 
\caption{Throughput vs training episodes for OMA and NOMA} 
\label{Fig.OMA-NOMA} 
\end{figure}

Fig. \ref{Fig.OMA-NOMA} shows the throughput versus the number of training episodes. It demonstrates the convergence and learning performance of the proposed MDQN algorithm for both the NOMA case and the OMA case. Since the parameters of the neural networks are initialized randomly and the $\epsilon$ is large at the beginning, the throughput of all cases is quite low and similar at inchoate episodes. At the first 70 episodes, there is almost no throughput increases in any curve, not only because of the large random action probability but also the replay memory buffer is being filled and the training will not start until the replay memory buffer is replete.

When the learning rate is 0.1, the excessive learning rate makes the NN unstable and can only obtain a small gain than random action choice. However, it is worth noting that the OMA scheme performs better than the NOMA scheme in this case, which indicates that the OMA scheme has a better tolerance than NOMA on unadvisable action selections. In the well-trained cases, when the learning rate is 0.01 or 0.001, the NOMA scheme outperforms the OMA scheme in terms of throughput. In the case of $lr = 0.001$, the data rate of the NOMA case exceeds the OMA case by $23\%$.

\begin{figure}[t!] 
\centering 
\includegraphics[width=0.5\textwidth]{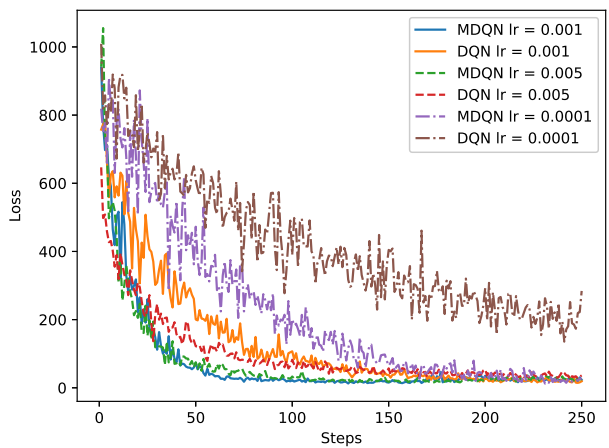} 
\caption{MDQN/DQN Loss vs training steps} 
\label{Fig.Loss} 
\end{figure}

Fig. \ref{Fig.Loss} compares the convergence rate of the MDQN and conventional DQN algorithm by plotting the loss described in \eqref{Loss}. It can be observed that the proposed MDQN paradigm has a higher training efficiency compared with the conventional independent agent mode. In this simulation, three UAVs connect to one NN via state abstraction as expounded in \textbf{Remark \ref{Expshare}}. As a consequence, it can be seen from the 3 pairs of curves, the number of training steps required by the DQN algorithm is approximately three times of the MDQN algorithm. To explain in another perspective, independent agent mode needs to train multiple NNs independently. When the agents have the same physical attribute, this scheme creates redundant training since these NNs are actually with high logical similarity. The proposed MDQN algorithm avoids this waste and thereby improves training efficiency.

\begin{figure}[t!]  
\centering 
\includegraphics[width=0.5\textwidth]{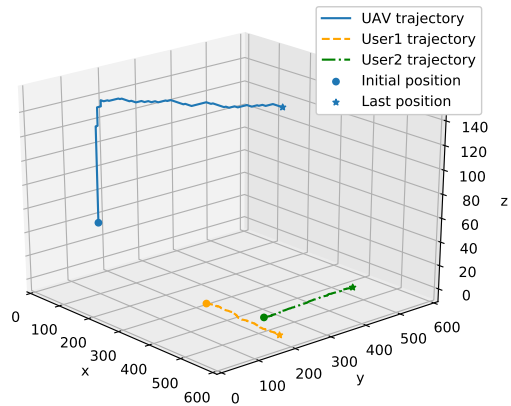} 
\caption{Optimized UAV trajectory in single UAV case} 
\label{Fig.5trajectory} 
\end{figure}

Fig. \ref{Fig.5trajectory} exhibits a UAV trajectory derived from the proposed MDQN algorithm as well as the users' movements. The users are following the directional mobility model and the duration time is 180s. It can be observed from the overall movement trend, UAV first chooses to increase the altitude, and then approach towards the user on a horizontal plane. Moreover, different from pre-set trajectories or the goal-oriented trajectory in \cite{sencing.RL}, this trajectory is more tortuous because it considers the data rate of each step during the movement.

\begin{figure}[t!] 
\centering 
\includegraphics[width=0.5\textwidth]{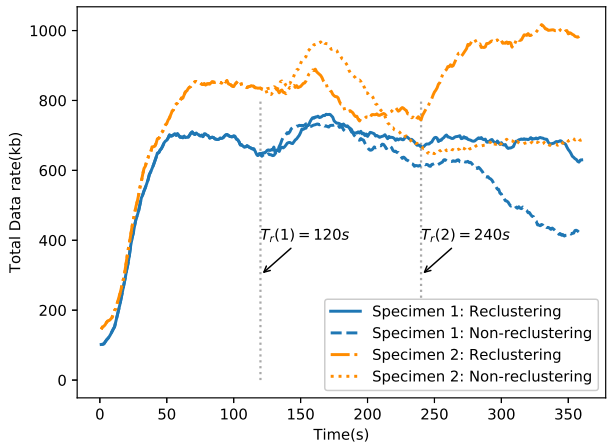} 
\caption{Data rate in test episode with/without re-clustering} 
\label{Fig.re-cls} 
\end{figure}

Fig. \ref{Fig.re-cls} shows the data rate of two specimens of both considering and without considering re-clustering in the test episode to reveal the role and value of re-clustering. In these simulations, three UAVs are employed, and the users follow the directional movement. The same model and parameters are set up in the two shown specimens, but the users have different initial distributions and directions of movement. In both specimens, since the initial user partition is the same, the obtained performances do not show difference in the first time interval $0\leq t\leq T_r,T_r = 120$. During this period of the service, the data rate rises rapidly and then stays at optimum since UAVs are well-trained and always take the optimal action. After the first re-clustering, non-re-clustered cases obtained similar or even prettier performances than re-clustered cases, but after the peak, the data rate of users without re-clustering receives a sustained decrease which does not appear in the re-clustered case. After the second re-clustering happened at $t=240$, the data rate gaps in two specimens become larger, where the re-clustered cases achieved significant advantages compared to the cases without re-clustering. The data rate diminution of the case without re-clustering can be ascribed to the lack of re-clustering since other conditions are exactly the same. This phenomenon suggests that in long-term service, re-clustering is beneficial to the data rate and it also provides the evidence for the insights in \textbf{Remark \ref{mobility}}.


\begin{figure}[t!] 
\centering 
\includegraphics[width=0.5\textwidth]{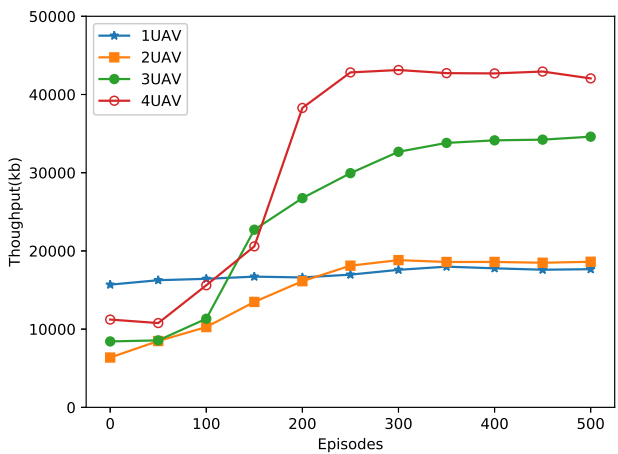} 
\caption{Throughput vs training episodes for different fleet size} 
\label{Fig.DifferentUAV} 
\end{figure}

Fig. \ref{Fig.DifferentUAV} characterizes the sum data rate over different UAV numbers and generally employing more UAVs can obtain larger throughput. In the case of single UAV, the throughput only obtains a limited increase over the training episodes due to the absence of inter-cluster interference and that is also the reason it even has similar performance with deploying 2 UAVs. As for efficiency, the gain obtained by adding the fourth UAV is less than adding the third one. Moreover, if the offloaded users are not completely blocked and are able to obtain some data traffic from GBS, a reasonable throughput compensation is supposed to be considered for the fewer UAV cases. Admittedly, more UAVs can provide access to more users but it can be claimed that the data rate efficiency of the UAV decreases as the fleet size increases.

\begin{figure}[t!] 
\centering 
\includegraphics[width=0.5\textwidth]{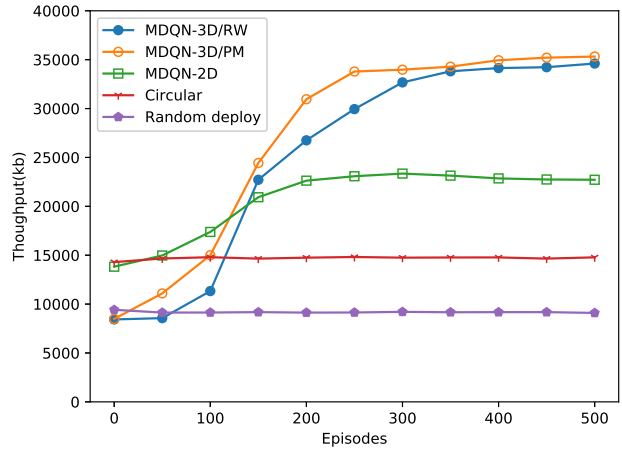} 
\caption{Throughput vs training episodes for different trajectory design scheme} 
\label{Fig.Deployment} 
\end{figure}

Fig. \ref{Fig.Deployment} compares the trajectory derived from the proposed MDQN algorithm with the benchmarks derived from the state-of-the-art. We verify the performance of the proposed method by invoking two mentioned user mobility models to prove its universality and the benchmarks are only simulated with random roaming users. The radius of the circular track is 150m, and the altitude is set to the empirical optimum. The well trained 3-D MDQN-derived trajectories are capable to achieve significant advantages over the 2-D trajectory and the circular trajectory. Compared to chaotic deployment, the circular trajectory has a better performance but inferior to all MDQN-derived trajectories.

\begin{figure}[t!] 
\centering 
\includegraphics[width=0.5\textwidth]{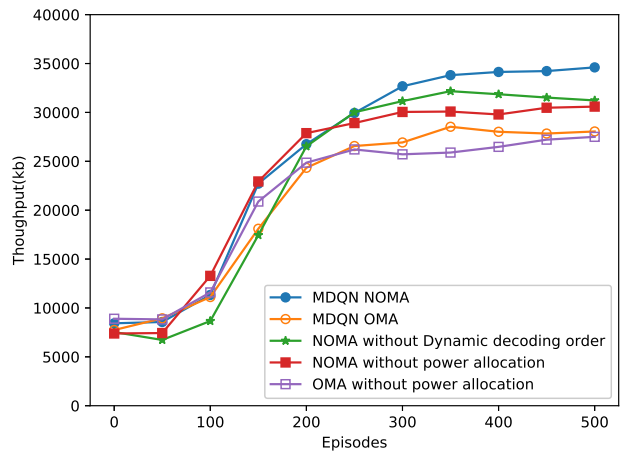} 
\caption{Throughput improvements of dynamic decoding order and power allocation } 
\label{Fig.Power-Decodingorder} 
\end{figure}

Fig.\ref{Fig.Power-Decodingorder} plots the contribution of power allocation and dynamic decoding order on the throughput in both NOMA and OMA cases. In cases where power allocation is absent, UAVs are assumed to transmit at maximum power. When the OMA scheme is adopted, RL-based power allocation has no absolute advantage over the maximum transmit power. It can be observed that in the NOMA case, the dynamic decoding order and the power allocation policy derived from the proposed MDQN algorithm are capable of achieving gains of approximately $12\%$ and $14\%$, respectively.

\section{Conclusions}

This paper was undertaken to design an effective paradigm for employing NOMA enhanced UAVs to assist terrestrial base stations and evaluated the performance of the proposed RL algorithm. An online machine learning based solution for tackling the formulated problem was proposed as well, where user clustering was determined by the K-means algorithm, 3-D deployments and power allocation were jointly optimized by the MDQN algorithm to maximize the total data rate of fleet-served users. Our simulation evaluated the performance of the proposed approach from multiple dimensions, including convergence, trajectory, multiple access schemes, fleet size and learning rate through the numerical results. These results proved the superiority of the NOMA UACO framework and the proposed MDQN paradigm possesses better convergence than the conventional DQN paradigm.

\bibliographystyle{IEEEtran}
\bibliography{UAVoffloading-arxiv}

\begin{thebibliography}{10}
\providecommand{\url}[1]{#1}
\csname url@samestyle\endcsname
\providecommand{\newblock}{\relax}
\providecommand{\bibinfo}[2]{#2}
\providecommand{\BIBentrySTDinterwordspacing}{\spaceskip=0pt\relax}
\providecommand{\BIBentryALTinterwordstretchfactor}{4}
\providecommand{\BIBentryALTinterwordspacing}{\spaceskip=\fontdimen2\font plus
\BIBentryALTinterwordstretchfactor\fontdimen3\font minus
  \fontdimen4\font\relax}
\providecommand{\BIBforeignlanguage}[2]{{%
\expandafter\ifx\csname l@#1\endcsname\relax
\typeout{** WARNING: IEEEtran.bst: No hyphenation pattern has been}%
\typeout{** loaded for the language `#1'. Using the pattern for}%
\typeout{** the default language instead.}%
\else
\language=\csname l@#1\endcsname
\fi
#2}}
\providecommand{\BIBdecl}{\relax}
\BIBdecl

\bibitem{UAVov.Zeng}
Y.~{Zeng}, R.~{Zhang}, and T.~J. {Lim}, ``Wireless communications with unmanned
  aerial vehicles: opportunities and challenges,'' \emph{IEEE Commun. Mag.},
  vol.~54, no.~5, pp. 36--42, May 2016.

\bibitem{Disasters.Zhao}
N.~{Zhao}, W.~{Lu}, M.~{Sheng}, Y.~{Chen}, J.~{Tang}, F.~R. {Yu}, and
  K.~{Wong}, ``{UAV}-assisted emergency networks in disasters,'' \emph{IEEE
  Wireless Commun.}, vol.~26, no.~1, pp. 45--51, February 2019.

\bibitem{Yuanwei.1}
Y.~{Liu}, Z.~{Qin}, Y.~{Cai}, Y.~{Gao}, G.~Y. {Li}, and A.~{Nallanathan},
  ``{UAV} communications based on non-orthogonal multiple access,'' \emph{IEEE
  Wireless Commun.}, vol.~26, no.~1, pp. 52--57, 2019.

\bibitem{Crowd}
Z.~{Wang}, L.~{Duan}, and R.~{Zhang}, ``Adaptive deployment for {UAV}-aided
  communication networks,'' \emph{IEEE Trans. Wireless Commun.}, vol.~18,
  no.~9, pp. 4531--4543, 2019.

\bibitem{JPA.Liu}
X.~{Liu}, J.~{Wang}, N.~{Zhao}, Y.~{Chen}, S.~{Zhang}, Z.~{Ding}, and F.~R.
  {Yu}, ``Placement and power allocation for {NOMA-UAV} networks,'' \emph{IEEE
  Wireless Commun. Lett.}, vol.~8, no.~3, pp. 965--968, June 2019.

\bibitem{U.distribution}
B.~{Galkin}, J.~{Kibilda}, and L.~A. {DaSilva}, ``Deployment of {UAV}-mounted
  access points according to spatial user locations in two-tier cellular
  networks,'' in \emph{2016 Wireless Days}, 2016, pp. 1--6.

\bibitem{SmallBS}
N.~{Bhushan}, J.~{Li}, D.~{Malladi}, R.~{Gilmore}, D.~{Brenner},
  A.~{Damnjanovic}, R.~T. {Sukhavasi}, C.~{Patel}, and S.~{Geirhofer},
  ``Network densification: the dominant theme for wireless evolution into
  {5G},'' \emph{IEEE Commun. Mag.}, vol.~52, no.~2, pp. 82--89, 2014.

\bibitem{UAVCellularMW}
Z.~{Xiao}, P.~{Xia}, and X.~{Xia}, ``Enabling {UAV} cellular with
  millimeter-wave communication: potentials and approaches,'' \emph{IEEE
  Commun. Mag.}, vol.~54, no.~5, pp. 66--73, 2016.

\bibitem{Edge.Cheng}
F.~{Cheng}, S.~{Zhang}, Z.~{Li}, Y.~{Chen}, N.~{Zhao}, F.~R. {Yu}, and V.~C.~M.
  {Leung}, ``{UAV} trajectory optimization for data offloading at the edge of
  multiple cells,'' \emph{IEEE Trans. Veh. Technol.}, vol.~67, no.~7, pp.
  6732--6736, July 2018.

\bibitem{offloading.Lyu}
J.~{Lyu}, Y.~{Zeng}, and R.~{Zhang}, ``{UAV}-aided offloading for cellular
  hotspot,'' \emph{IEEE Trans. Wireless Commun.}, vol.~17, no.~6, pp.
  3988--4001, 2018.

\bibitem{Cyclical.Lyu}
J.~Lyu, Y.~Zeng, and R.~Zhang, ``Spectrum sharing and cyclical multiple access
  in {UAV}-aided cellular offloading,'' in \emph{IEEE GLOBECOM 2017}.\hskip 1em
  plus 0.5em minus 0.4em\relax IEEE, 2017, pp. 1--6.

\bibitem{PPP.Turgut}
E.~{Turgut} and M.~C. {Gursoy}, ``Downlink analysis in unmanned aerial vehicle
  {(UAV)} assisted cellular networks with clustered users,'' \emph{IEEE
  Access}, vol.~6, pp. 36\,313--36\,324, 2018.

\bibitem{Spectrum.Hu}
Z.~{Hu}, Z.~{Zheng}, L.~{Song}, T.~{Wang}, and X.~{Li}, ``{UAV} offloading:
  Spectrum trading contract design for {UAV}-assisted cellular networks,''
  \emph{IEEE Trans. Wireless Commun.}, vol.~17, no.~9, pp. 6093--6107, Sep.
  2018.

\bibitem{3D.Yaliniz}
R.~I. {Bor-Yaliniz}, A.~{El-Keyi}, and H.~{Yanikomeroglu}, ``Efficient {3-D}
  placement of an aerial base station in next generation cellular networks,''
  in \emph{IEEE ICC 2016}, May 2016, pp. 1--5.

\bibitem{SolarUAV}
Y.~{Sun}, D.~{Xu}, D.~W.~K. {Ng}, L.~{Dai}, and R.~{Schober}, ``Optimal
  {3D}-trajectory design and resource allocation for solar-powered {UAV}
  communication systems,'' \emph{IEEE Trans. Commun.}, vol.~67, no.~6, pp.
  4281--4298, 2019.

\bibitem{NOMAoff.Kim}
P.~K. {Sharma} and D.~I. {Kim}, ``{UAV}-enabled downlink wireless system with
  non-orthogonal multiple access,'' in \emph{IEEE GC Wkshps 2017}, 2017, pp.
  1--6.

\bibitem{Hover.Song}
Q.~{Song}, F.~{Zheng}, and S.~{Jin}, ``Multiple {UAV}s enabled data offloading
  for cellular hotspots,'' in \emph{IEEE WCNC 2019}, 2019, pp. 1--6.

\bibitem{A2GNOMA.Mu}
X.~{Mu}, Y.~{Liu}, L.~{Guo}, and J.~{Lin}, ``Non-orthogonal multiple access for
  air-to-ground communication,'' \emph{IEEE Trans. Commun.}, pp. 1--1, 2020.

\bibitem{UAVDLNOMA.Lei}
L.~{Wang}, B.~{Hu}, S.~{Chen}, and J.~{Cui}, ``{UAV}-enabled reliable mobile
  relaying based on downlink {NOMA},'' \emph{IEEE Access}, vol.~8, pp.
  25\,237--25\,248, 2020.

\bibitem{liu2020}
X.~Liu, M.~Chen, Y.~Liu, Y.~Chen, S.~Cui, and L.~Hanzo, ``Artificial
  intelligence aided next-generation networks relying on {UAV}s,'' \emph{arXiv
  preprint arXiv:2001.11958}, 2020.

\bibitem{TD.Xiao}
X.~{Liu}, Y.~{Liu}, Y.~{Chen}, and L.~{Hanzo}, ``Trajectory design and power
  control for multi-{UAV} assisted wireless networks: A machine learning
  approach,'' \emph{IEEE Trans. Veh. Technol.}, vol.~68, no.~8, pp. 7957--7969,
  2019.

\bibitem{Q.Lingyang}
J.~{Hu}, H.~{Zhang}, and L.~{Song}, ``Reinforcement learning for decentralized
  trajectory design in cellular {UAV} networks with sense-and-send protocol,''
  \emph{IEEE Internet Things J.}, vol.~6, no.~4, pp. 6177--6189, 2019.

\bibitem{MLOV.MAO}
Y.~{Sun}, M.~{Peng}, Y.~{Zhou}, Y.~{Huang}, and S.~{Mao}, ``Application of
  machine learning in wireless networks: Key techniques and open issues,''
  \emph{IEEE Commun. Surveys Tuts.}, vol.~21, no.~4, pp. 3072--3108, 2019.

\bibitem{DQN.IOT}
W.~{Liu}, P.~{Si}, E.~{Sun}, M.~{Li}, C.~{Fang}, and Y.~{Zhang}, ``Green
  mobility management in {UAV}-assisted {IoT} based on dueling {DQN},'' in
  \emph{IEEE ICC 2019}, 2019, pp. 1--6.

\bibitem{4ML.UA}
J.~{Hu}, H.~{Zhang}, L.~{Song}, Z.~{Han}, and H.~V. {Poor}, ``Reinforcement
  learning for a cellular internet of {UAV}s: Protocol design, trajectory
  control, and resource management,'' \emph{IEEE Wireless Commun.}, vol.~27,
  no.~1, pp. 116--123, 2020.

\bibitem{Paradigms}
C.~{Jiang}, H.~{Zhang}, Y.~{Ren}, Z.~{Han}, K.~{Chen}, and L.~{Hanzo},
  ``Machine learning paradigms for next-generation wireless networks,''
  \emph{IEEE Wireless Commun.}, vol.~24, no.~2, pp. 98--105, 2017.

\bibitem{MLWN.OV}
Y.~{Sun}, M.~{Peng}, Y.~{Zhou}, Y.~{Huang}, and S.~{Mao}, ``Application of
  machine learning in wireless networks: Key techniques and open issues,''
  \emph{IEEE Commun. Surveys Tuts.}, vol.~21, no.~4, pp. 3072--3108, 2019.

\bibitem{niknam2019federated}
S.~Niknam, H.~S. Dhillon, and J.~H. Reed, ``Federated learning for wireless
  communications: Motivation, opportunities and challenges,'' \emph{arXiv
  preprint arXiv:1908.06847}, 2019.

\bibitem{Local.OPT}
J.~M. {Rojas} and G.~{Fraser}, ``Is search-based unit test generation research
  stuck in a local optimum?'' in \emph{IEEE/ACM SBST 2017}, 2017, pp. 51--52.

\bibitem{NOMA5G.yuanwei}
Y.~{Liu}, Z.~{Qin}, M.~{Elkashlan}, Z.~{Ding}, A.~{Nallanathan}, and
  L.~{Hanzo}, ``Nonorthogonal multiple access for {5G} and beyond,''
  \emph{Proc. IEEE}, vol. 105, no.~12, pp. 2347--2381, 2017.

\bibitem{GBS.power}
F.~{Guidolin} and M.~{Nekovee}, ``Investigating spectrum sharing between {5G}
  millimeter wave networks and fixed satellite systems,'' in \emph{2015 IEEE GC
  Wkshps}, 2015, pp. 1--7.

\bibitem{3gpp.36.777}
3GPP, ``{Technical Specification Group Radio Access Network; Study on Enhanced
  LTE Support for Aerial Vehicles},'' {3rd Generation Partnership Project
  (3GPP)}, Technical Specification (TS) 36.777, 01 2018, version 15.0.0.

\bibitem{cui.signal}
J.~{Cui}, Y.~{Liu}, Z.~{Ding}, P.~{Fan}, and A.~{Nallanathan}, ``Optimal user
  scheduling and power allocation for millimeter wave {NOMA} systems,''
  \emph{IEEE Trans. Wireless Commun.}, vol.~17, no.~3, pp. 1502--1517, 2018.

\bibitem{SIC}
M.~M. {Alsmadi}, N.~{Abu Ali}, M.~{Hayajneh}, and S.~S. {Ikki}, ``Down-link
  {NOMA} networks in the presence of {IQI} and imperfect {SIC}: Receiver design
  and performance analysis,'' \emph{IEEE Trans. Veh. Technol.}, pp. 1--1, 2020.

\bibitem{NP.HARD}
S.~{Zhang}, H.~{Zhang}, B.~{Di}, and L.~{Song}, ``Cellular {UAV}-to-{X}
  communications: Design and optimization for multi-{UAV} networks,''
  \emph{IEEE Trans. on Wireless Commun.}, vol.~18, no.~2, pp. 1346--1359, 2019.

\bibitem{cui.clustering}
J.~{Cui}, Z.~{Ding}, P.~{Fan}, and N.~{Al-Dhahir}, ``Unsupervised machine
  learning-based user clustering in millimeter-wave-{NOMA} systems,''
  \emph{IEEE Trans. Wireless Commun.}, vol.~17, no.~11, pp. 7425--7440, 2018.

\bibitem{liu2019MDP}
L.~Liu, B.~Tian, X.~Zhao, and Q.~Zong, ``{UAV} autonomous trajectory planning
  in target tracking tasks via a {DQN} approach,'' in \emph{2019 IEEE
  International Conference on Real-time Computing and Robotics (RCAR)}.\hskip
  1em plus 0.5em minus 0.4em\relax IEEE, 2019, pp. 277--282.

\bibitem{chen2020bellmen}
Z.~Chen, Y.~Zhong, X.~Ge, and Y.~Ma, ``An actor-critic-based {UAV}-bss
  deployment method for dynamic environments,'' \emph{arXiv preprint
  arXiv:2002.00831}, 2020.

\bibitem{mnih2013playing}
V.~Mnih, K.~Kavukcuoglu, D.~Silver, A.~Graves, I.~Antonoglou, D.~Wierstra, and
  M.~Riedmiller, ``Playing atari with deep reinforcement learning,''
  \emph{arXiv preprint arXiv:1312.5602}, 2013.

\bibitem{sencing.RL}
J.~{Hu}, H.~{Zhang}, K.~{Bian}, L.~{Song}, and Z.~{Han}, ``Distributed
  trajectory design for cooperative internet of {UAV}s using deep reinforcement
  learning,'' in \emph{IEEE GLOBECOM 2019}, 2019, pp. 1--6.

\end{thebibliography}

\end{document}